\documentclass{article}
\usepackage{spconf,amsmath,graphicx}
\usepackage{algorithm}
\usepackage{algpseudocode}
\usepackage{comment}
\usepackage{soul}
\usepackage{svg}
\usepackage{amsmath}
\usepackage{enumitem}
\usepackage{xcolor}

\title{FLToP CTC: Frame-Level Token Pruning via Relative Threshold for Efficient and Memory-Saving Decoding on Diverse Platforms}
\name{Atul Shree, Harshith Jupuru}
\address{\{atul, harshith\}@convin.ai}
\begin{document}
%
\maketitle
\begin{abstract}
\label{sec:abstract}
CTC-based ASR systems face computational and memory bottlenecks in resource-limited environments. Traditional CTC decoders, requiring up to~90\% of processing time in systems (e.g., wav2vec2-large on L4 GPUs), face inefficiencies due to exhaustive token-level operations. This paper introduces \textbf{Frame Level Token Pruning for Connectionist Temporal Classification (FLToP CTC)}, a novel decoding algorithm that employs frame-level token pruning guided by a relative threshold probability. By dynamically eliminating low-probability tokens per frame, FLToP CTC reduces compute and memory demands while maintaining negligible WER degradation. On LibriSpeech, FLToP CTC achieves a~$10.5\times$ runtime speedup and~$2.78\times$ memory reduction versus standard CTC decoders. Its simplicity enables seamless integration into CTC decoders across platforms (CPUs, GPUs, etc.). FLToP CTC addresses CTC bottlenecks, offering scalability for resource-limited environments and realtime applications, enhancing speech recognition accessibility and efficiency.
\end{abstract}
\section{Introduction}
\label{sec:introduction}

\noindent Connectionist Temporal Classification (CTC) \cite{book,10.1145/1143844.1143891,pmlr-v32-graves14} is a widely adopted algorithm for automatic speech recognition (ASR) due to its alignment-free approach enabling end-to-end training and ability to handle variable-length sequences. CTC-based models, such as those integrated into popular architectures like wav2vec \cite{DBLP:journals/corr/abs-2006-11477} and wavLM \cite{Chen_2022}, have gained prominence for their ability to map speech inputs to text outputs without explicit time alignments. However, despite their effectiveness, CTC-based ASR systems face significant computational and memory bottlenecks during decoding, especially in resource-constrained environments. A key limitation of traditional CTC decoders is inefficient token processing. At every step, these evaluate all possible tokens, leading to significant computational costs and memory usage. This becomes particularly challenging in large models, where CTC decoding (which runs on CPUs) can account for as much as~90\% of the processing time, even on systems equipped with L4 GPUs that are paired with wav2vec large encoders. The issue is further amplified in real-time scenarios and low-resource devices, where restricted computational power and memory limitations necessitate more streamlined approaches. This paper introduces \textbf{FLToP CTC} (Frame-Level Token Pruning for CTC), a novel decoding algorithm that addresses these bottlenecks by leveraging dynamic frame-level token pruning driven by a relative threshold probability of the top token. Instead of exhaustively processing all tokens at each frame, FLToP CTC dynamically eliminates low-probability candidates, reducing computational and memory demands. This approach maintains negligible word error rate (WER) degradation while achieving significant efficiency gains. In a nutshell, we make the following contributions:
\begin{itemize}[nosep,noitemsep,leftmargin=*]
    \item \textbf{Dynamic Frame-Level Pruning Mechanism}: We introduce new decoding algorithm \textbf{FLToP CTC}. The algorithm introduces a dynamic pruning mechanism that operates at the frame level to retain only high-confidence tokens.
    \item \textbf{Platform-Agnostic Algorithm Design}: The algorithm is designed to be simple, and versatile, enabling seamless integration into existing CTC decoders across diverse platforms, including CPUs, GPUs, and low-resource hardware.
    \item \textbf{Empirical Validation Through Statistical Behavior Study}: We present a thorough study of CTC decoder statistics and behaviors, which underpins the design of FLToP CTC and validates its effectiveness across various deployment settings.
\end{itemize}

\section{Related Work}
\label{sec:related_work}

State-of-the-art CTC decoders, such as those integrated with KenLM \cite{heafield-2011-kenlm} and Flashlight \cite{kahn2022flashlight}, employ static top-N pruning to prioritize likely hypotheses. While effective, static pruning lacks frame-level adaptivity, causing redundant computation in “easy” frames where stronger pruning could reduce latency without harming accuracy. Prior efforts to optimize CTC decoders for low-resource hardware include model compression, beam search refinements, and memory-efficient implementations \cite{Lu_2019,10.1109/TVLSI.2017.2717950,wang2018clstmenablingefficientlstm}, but these too rely on static pruning. GPU-specific accelerations \cite{galvez2023gpuacceleratedwfstbeamsearch} improve speed but neglect CPU and constrained-device scenarios. Work on transducer (RNN-T) models \cite{galvez24_interspeech,kuang22_interspeech,tian2021fsracceleratinginferenceprocess,wang2022acceleratingrnnttraininginference} provides insights but is not directly transferable to CTC due to architectural and algorithmic differences. Other strategies, such as nucleus sampling \cite{holtzman2020curiouscaseneuraltext}, accumulate tokens until a threshold is met, often introducing redundant candidates.

Reducing search space via a universal, language-independent character set \cite{verma2023asr} has shown promise for multilingual ASR, limiting vocabulary sizes. Our approach is applicable to such systems as well by introducing dynamic frame-level pruning, focusing computation on relevant tokens while preserving model capability.

Traditional toolkits like Kaldi \cite{povey2011kaldi} and HARPY \cite{lowerre1990harpy} employ HMM/Viterbi-based state-dependent pruning, combining acoustic and transition probabilities at a higher granularity. In contrast, we prune tokens dynamically at the frame level, independent of evolving hypotheses, enabling more adaptive and fine-grained control. This makes our method particularly effective for CTC-based models such as WavLM and Wav2Vec2, which output frame-level emissions.

Overall, our frame-level pruning strategy complements existing techniques, enhances candidate selection, and is easily adaptable across diverse ASR frameworks.
\section{FLToP CTC: Idea and Algorithm}
\label{sec:idea_and_algorithm}

\begin{algorithm}
\caption{Beam Search FLToP CTC Decoding for ASR}
\label{algo:beam_search_fltop_ctc_decoding_for_asr}
\begin{algorithmic}[1]
\Procedure{BeamSearchFLToPCTC}{\textit{logits}, \textit{beam\_size}, \textit{beam\_threshold}, \textit{LM}, \textit{N}, \textit{R}}
    \State $\textit{B} \gets \{(\epsilon, 0)\}$
    \For{\textit{t} in $0 \ldots T$}
        \State $\textit{B'} \gets \{\}$
        \State $\textit{logits\_idx\_sorted} \gets \text{PartialSortDesc}(\textit{logits[t]}, \textit{N})$
        \State $\textit{logit\textsubscript{t0}} \gets \textit{logits}[t][\textit{logits\_idx\_sorted}[0]]$
        \For{\textit{(prefix, score)} in \textit{B}}
            \For{\textit{i} in  $0 \ldots N$}
                \State $\textit{logit\textsubscript{ti}} \gets \textit{logits}[t][\textit{logits\_idx\_sorted}[i]]$
                \If{$\textit{logit\textsubscript{ti}} \leq \textit{logit\textsubscript{t0}} * R$}
                    \State \textbf{break}
                \EndIf
                \State $\textit{token} \gets \text{IdToToken}(\textit{logits\_idx\_sorted}[i])$
                \State $\textit{prefix'} \gets \textit{prefix} + \textit{token}$
                \State $\textit{score'} \gets \textit{score} + \textit{logit\textsubscript{ti}}$
                \State $\textit{score'} \gets \textit{score'} + \text{LM}(\textit{prefix'})$
                \State $\textit{B'}.\text{add}((\textit{prefix'}, \textit{score'}))$
            \EndFor
        \EndFor
        \State $\textit{B} \gets \text{SelectTopK}(\textit{B'}, \textit{beam\_size}, \textit{beam\_threshold})$
    \EndFor
    \State \Return $\text{GetHighestScorePrefix}(\textit{B})$
\EndProcedure
\end{algorithmic}
\end{algorithm}

\graphicspath{{./images_pdf/}}
\begin{figure}[t]
    \centering
    \includegraphics[width=0.50\textwidth]{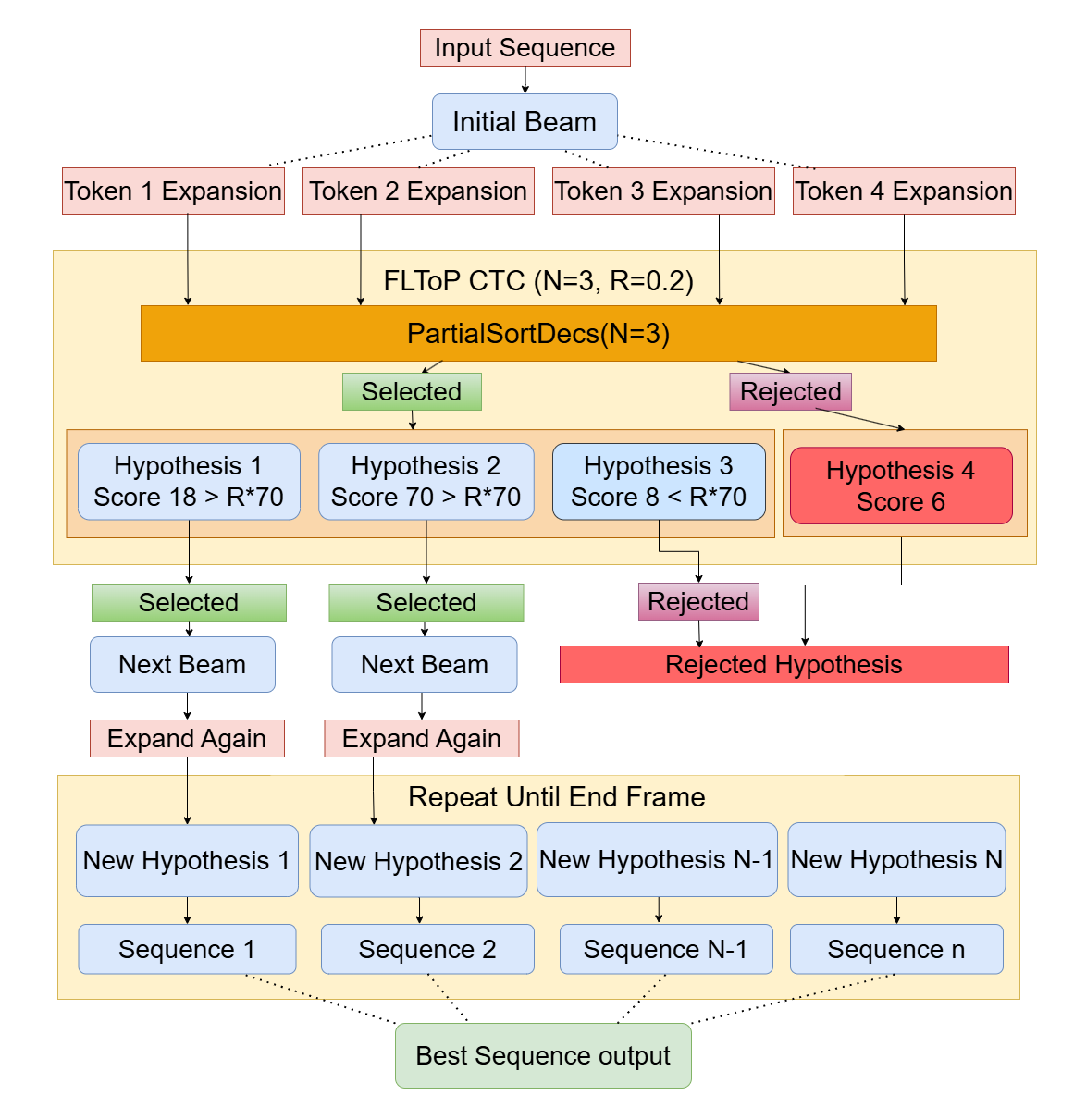}
    \vspace{-7mm}
    \caption{Workflow of FLToP CTC Algorithm}
    \vspace{-5mm}
    \label{fig:algo-diagram}
\end{figure}

\noindent To address inefficiencies in standard BeamSearchCTC decoders, we propose Frame-Level Token Pruning for CTC (FLToP CTC) [Algorithm~\ref{algo:beam_search_fltop_ctc_decoding_for_asr}, Fig.~\ref{fig:algo-diagram}]. The algorithm first selects the top-N tokens from current frame (lines 5 \& 8), then applies a secondary pruning step that retains only tokens with scores above a relative threshold $R$ times the highest score (lines 10–12), a hyperparameter that adjusts pruning intensity to optimize computational resources. This two-stage process eliminates low-probability candidates while focusing computation on the most promising ones.

The novelty lies in the conditional break (lines 10–12), which makes FLToP simple, generic, and platform-independent, enabling integration across CPUs, GPUs, and other environments. Fig.~\ref{fig:algo-diagram} illustrates the workflow: an initial beam of hypotheses (line 2) expands with top-N tokens (line 13–17), prunes via threshold $R$ (e.g., $R=0.2$ in Fig.~\ref{fig:algo-diagram}), and retains top candidates (line 20) for the next timestep. This iterative process continues until the audio ends, after which the highest-scoring sequence is returned (line 22).

\section{Experiments and Results}
\label{sec:experiments}

\noindent All experiments related to ASR CTC decoding were conducted utilizing the well known LibriSpeech dataset \cite{7178964} for easy reproducibility of the results. The training set served to develop a 4-gram KenLM-based language model (LM) utilized in CTC decoding, as well as to compile the vocabulary and lexicon files. Evaluations were carried out using the dev-clean, dev-other, test-clean, and test-other subsets of LibriSpeech. The encoding of audio was performed using the wav2vec-2 large model, which was trained on Librispeech and LibriVox \cite{pratap2020mls} data and fine-tuned with 960 hours of Librispeech dataset. For the CTC decoding experiments, tools such as flashlight-text, fairseq \cite{ott2019fairseq}, and KenLM were employed.

\noindent To conduct a thorough evaluation of different CTC decoding strategies, we established a specific configuration setup. Our vocabulary includes 32 tokens: the 26 letters of the English alphabet, an apostrophe, a space character ($|$), along with bos, pad, eos, and unk. We set lm-weight=1, sil-score=0.0, word-score=0.95, beam-threshold=25, with all 32 tokens, and 1000 beam-size. We refer to this arrangement as the \textbf{Baseline} Configuration in this paper.

\noindent We tested two pruning strategies: \textbf{Top-N}, limiting the search to top-N tokens, and \textbf{FLToP CTC}, our proposed method integrating relative token threshold pruning with top-N selection.

\subsection{Index tracking of Chosen Token} \label{subsec:exp_index_track_chosen_token}
\begin{figure}[t]
    \centering
    \includegraphics[width=0.47\textwidth]{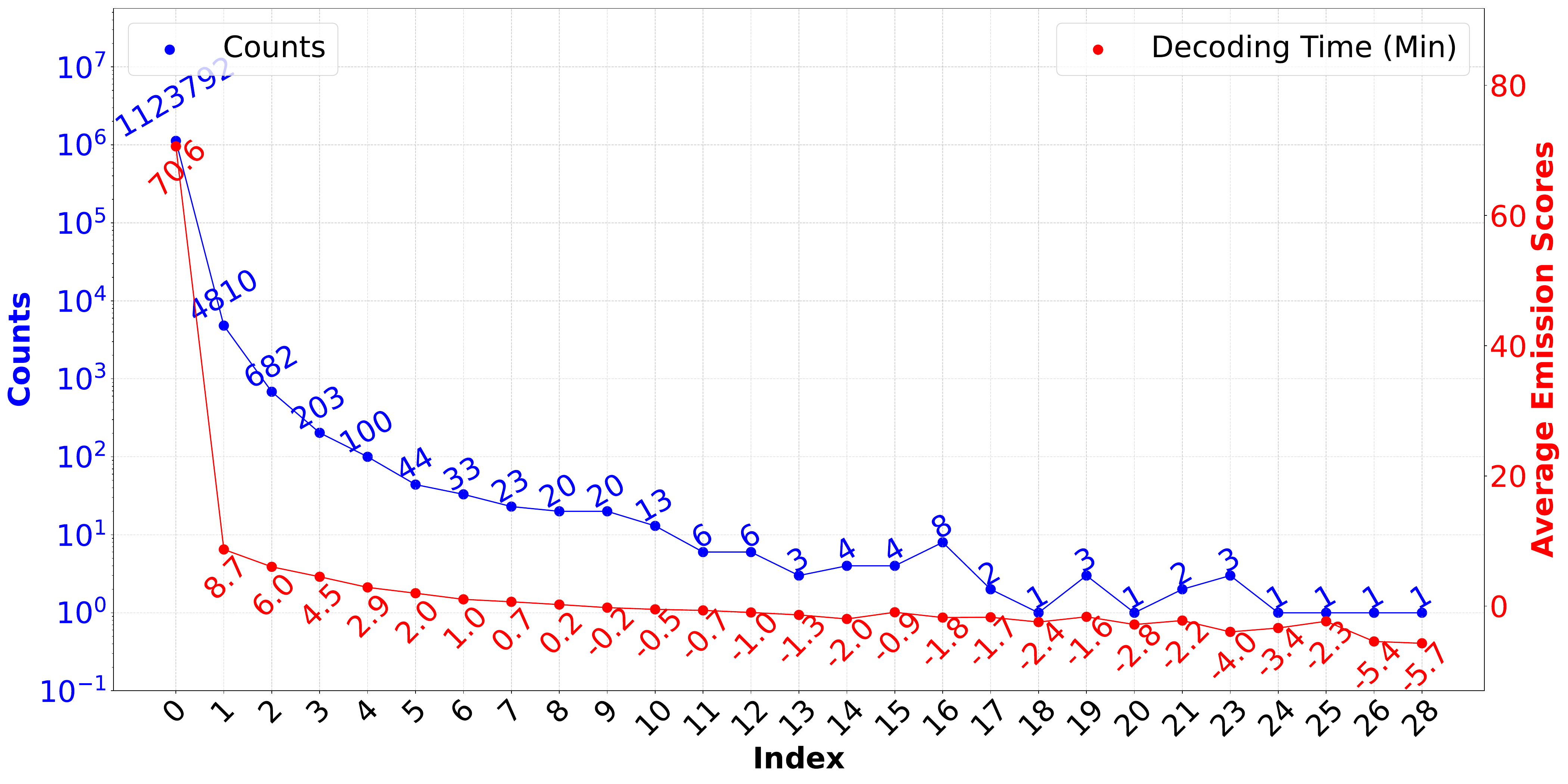}
    \vspace{-4mm}
    \caption{Count and Average Emission Scores of choosing a token at specific index (from best beam from all test samples) after sorting the token based on emission scores}
     \vspace{-4mm}
\label{fig:resultsindex_counts_stats_sortedTokenProbs_batchsize32_softmax_numHyps_tokenSize32_tokenRel_0.0}
\end{figure}

For this experiment, we employ the Baseline Configuration as defined earlier. During the beam search process, we monitor the sorted indices and corresponding emission scores of selected tokens for each candidate hypothesis at every expansion step. This monitoring allows us to evaluate the frequency with which tokens are chosen at different indices throughout the entire dataset and across all timesteps. Analyzing this distribution helps us identify the optimal number of top N tokens necessary for adequate token coverage when implementing FLToP CTC or TopN pruning strategies. While the specific tokens at each index may change across timesteps, our method records how frequently tokens are selected at each position. This data is vital for adjusting the TopN and relative-token-threshold parameters, optimizing both accuracy and efficiency in FLToP CTC decoding. As shown in Figure \ref{fig:resultsindex_counts_stats_sortedTokenProbs_batchsize32_softmax_numHyps_tokenSize32_tokenRel_0.0}, the algorithm predominantly selects the top 1-4 tokens with $99.9823\%$ of the time supporting the strategy of retaining only Top-4 tokens for beam expansion. 
\noindent Furthermore, the plot in Figure~\ref{fig:resultsindex_counts_stats_sortedTokenProbs_batchsize32_softmax_numHyps_tokenSize32_tokenRel_0.0} shows the average emission scores of tokens chosen at each index which gives an idea of token significance at each index position.


\graphicspath{{./images_pdf/}}



\subsection{Top-N thresholding for N = 1 \ldots 32 tokens}
\label{subsec:exp_top_1_to_32}

\begin{figure}[t]
    \centering
    \includegraphics[width=0.47\textwidth]{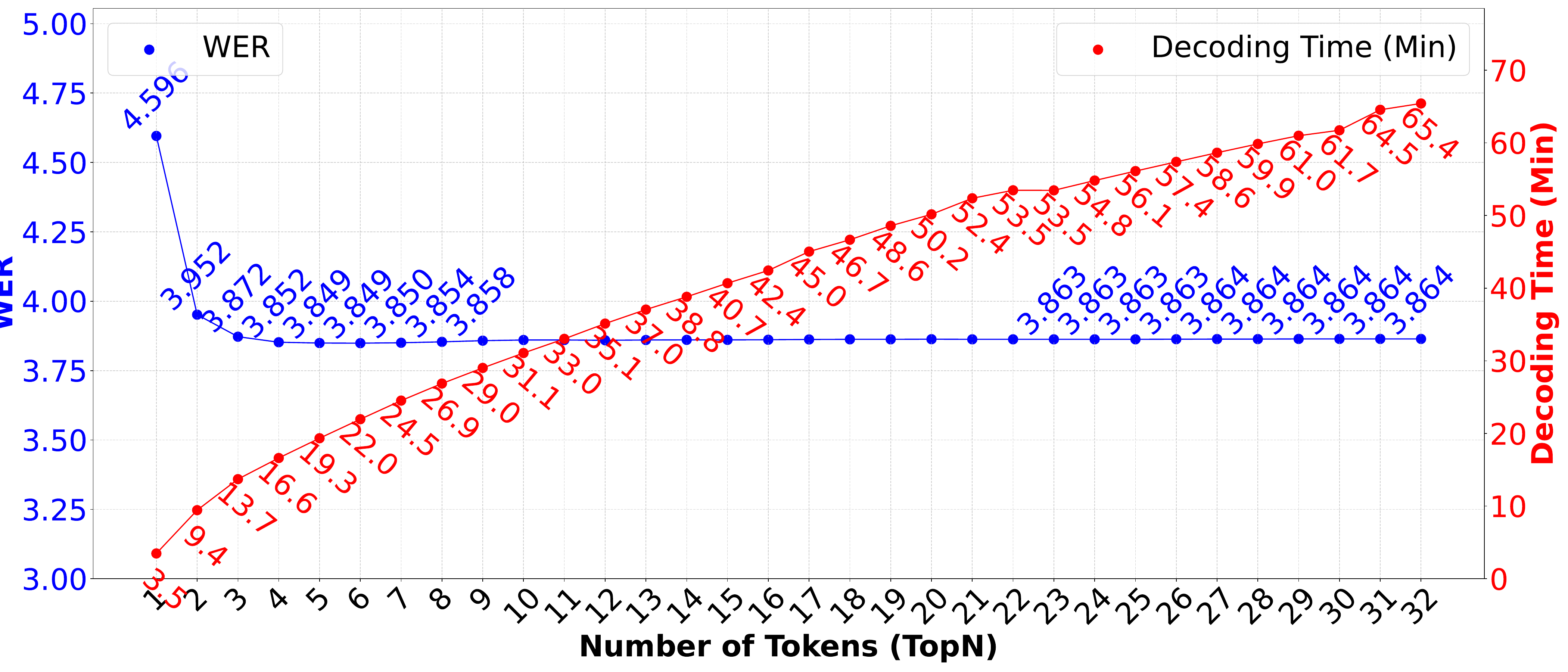}
    \caption{WER and Time Taken for Decoding by varying the beam size token}
\label{fig:wer_vs_decoding_time_relTokThres0.0_varyNumTokens_side_by_side_scatter}
\end{figure}

This experiment explores the effects of altering the beam-size-token parameter, which dictates the number of top-ranked tokens included during beam search. We assess configurations from Top-1, where only the highest-scoring token per timestep is considered, to Top-32, which includes all tokens and is analogous to the baseline configuration. As the number of tokens increases, the search space widens, potentially improving transcription accuracy but also increasing decoding time. However, as shown in Figure~\ref{fig:wer_vs_decoding_time_relTokThres0.0_varyNumTokens_side_by_side_scatter}, enhancements in the WER become minimal beyond $N=4$. Notably, limiting the search to Top-4 tokens results in a WER of 3.852, surpassing the baseline's exhaustive search (with Top-32, WER = 3.864). Additionally, decoding time almost linearly increases with the number of beam-size-tokens. The baseline setting (Top-32) experiences a decoding time that is 3.94× longer than the Top-4 configuration, without any WER benefits which supports to use Top4 tokens only to balance both accuracy and computational demands. Based on these findings, we set N=4 for subsequent experiments to further enhance WER while minimizing computational requirements.


\graphicspath{{./images_pdf/}}


\subsection{Relative Token Thresholding (N = 4, R varying)}
\label{subsec:exp_rel_token_thres_top4}
\begin{figure}[t]
    \centering
    \includegraphics[width=0.47\textwidth]{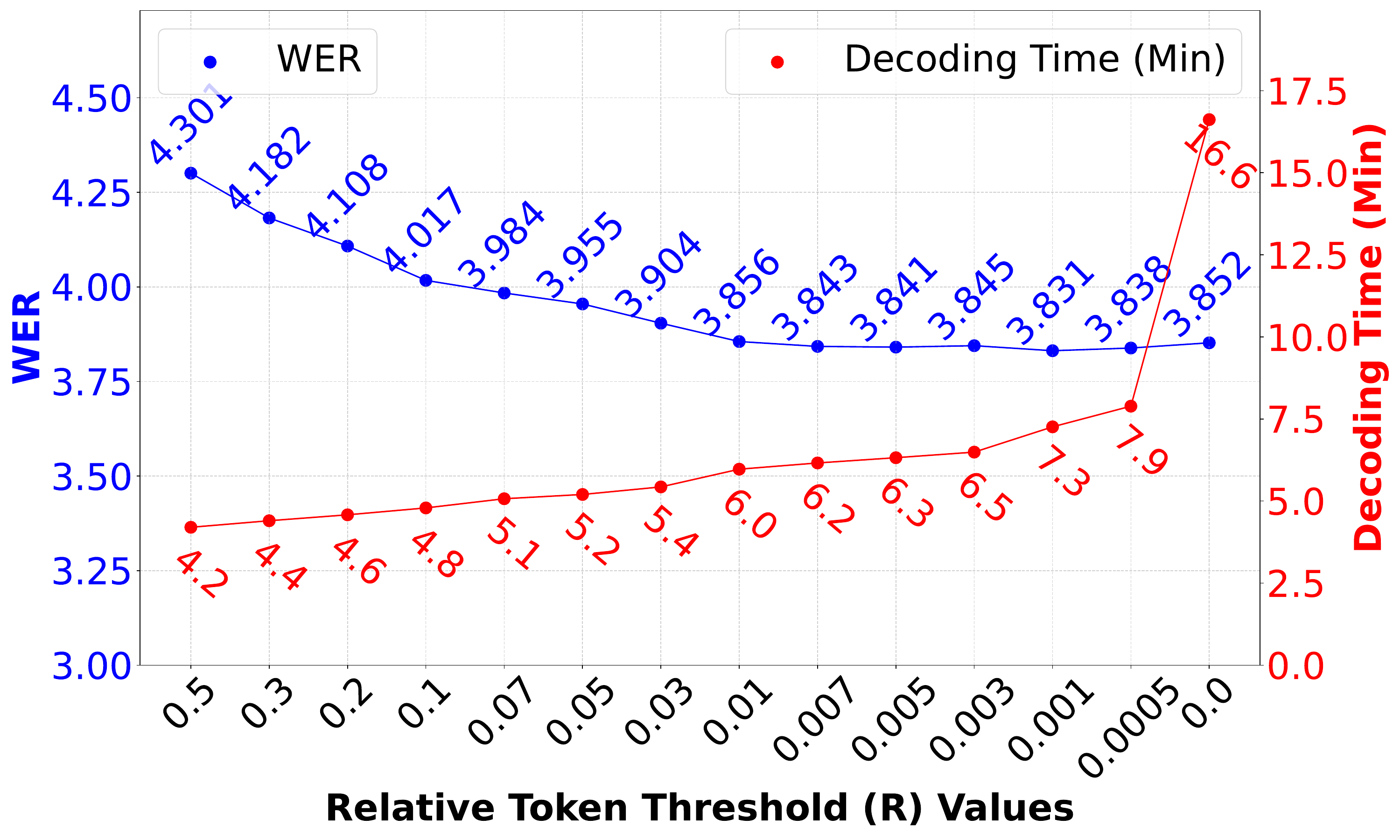}
    \vspace{-2mm}
    \caption{WER and Time Taken for Decoding by varying the Relative Token Threshold with Top4 approach}
    \vspace{-5mm}
    \label{fig:wer_vs_decoding_time_top4_varyRelTokThres_side_by_side_scatter}
\end{figure}

Building on previous results, we found that limiting the beam-size-token of 4 achieves a WER comparable to the baseline configuration. To further boost computational efficiency, we adjusted the parameter R in Algorithm \ref{algo:beam_search_fltop_ctc_decoding_for_asr} to explore the trade-off between WER and computational costs, while keeping the beam size and other decoding parameters constant. By selectively pruning tokens at each timestep, our goal was to minimize unnecessary expansion of the search space while maintaining transcription accuracy. As demonstrated in Figure~\ref{fig:wer_vs_decoding_time_top4_varyRelTokThres_side_by_side_scatter}, setting R to 0.007 resulted in the same WER but reduced the decoding time to 369.6 seconds, achieving a speed~$2.78\times$ faster than the Top-4 method and~$10.5\times$ faster than the baseline. Furthermore, WER slightly improves to 3.843 from 3.852, which is seen in the Top-4 method without relative pruning. The optimal WER recorded is 3.831 at $R= 0.001$, which processes faster than using Top-4 pruning alone, although it is marginally slower than at the 0.007 threshold due to reduced pruning.

Despite the wide range of $R$, which adjusts the pruning effect significantly (allowing tokens whose emission score is at $R \times$ of the top token's score), the time difference is minimal with R starting from $0.5$ to another extreme of $0.03$. Typically, only 1 to 2 tokens from the top 4 are selected for beam search with both $0.5$ and $0.003$ as $R$ values. However, with larger vocabularies and higher Top-N settings, the benefits of relative token pruning become more evident. With more tokens eligible for pruning, significant improvements in decoding speed can be achieved, especially when the encoder outputs are highly confident. When emission scores are more evenly distributed across tokens, varying $R$ affects decoding performance more substantially.


\graphicspath{{./images_pdf/}}


\subsection{Relative Token Thresholding impact on Number of Beams and Memory Consumption} 
\label{subsec:exp_rel_tok_impact_on_memory}
\begin{figure}[t]
    \centering
    \includegraphics[width=0.50\textwidth]{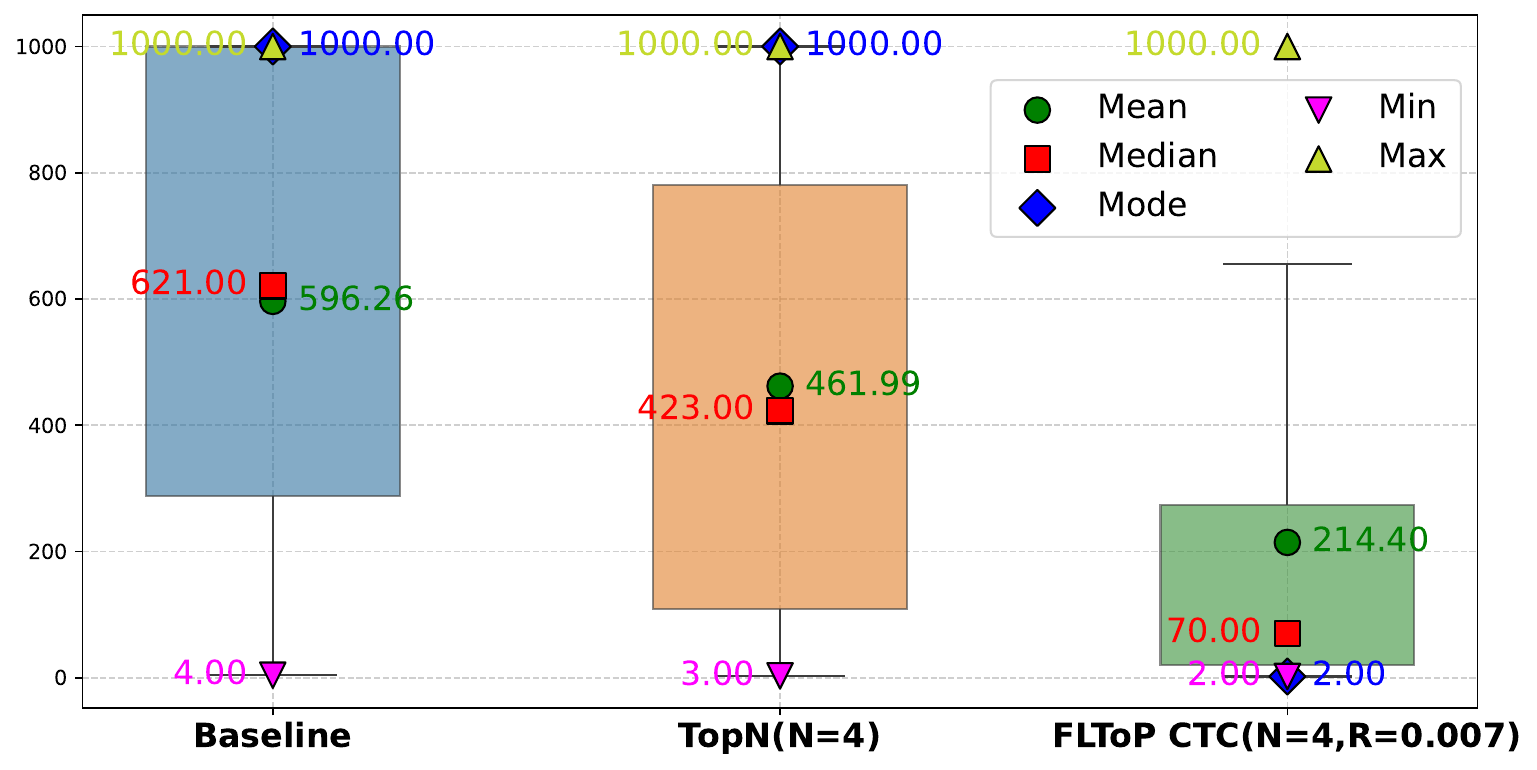}
    \caption{Box plot for Overall Number of candidates stored in beam search for all time steps across all test samples}
    \label{fig:results_hypothesis_stats_top4_relTokThres0.007}
\end{figure}

This study evaluates the effect of token pruning on beam count throughout various decoding strategies. We monitored the number of hypotheses at each timestep across the dataset under three different setups: the Baseline Configuration, which considers all tokens in the beam search; Top-N Pruning with N=4, limiting the search to the top four tokens; and FLToP CTC, our proposed method using $N=4$ and $R=0.007$, which showed enhanced performance in prior tests. By tracking the number of beams during decoding, we gain insights into the memory and computational demands of each method. A reduction in average beam count directly results in lower memory usage and faster decoding speeds. This experiment aims to quantify the efficiency improvements from relative token pruning in minimizing the search space while preserving transcription accuracy. According to Figure~\ref{fig:results_hypothesis_stats_top4_relTokThres0.007}, FLToP CTC with settings [N=4, R=0.007] achieves approximately ~$2.78\times$ fewer beams on average.

On average, our approach maintains 214.4 beams, whereas the baseline configuration and the Top-N pruning (N=4) method maintain 596.26 beams ($2.78 \times$ more) and 461.99 beams ($2.15 \times$ more), respectively. The box plot further highlights that the mean, median, and quartiles for beam counts in our method are consistently lower than those in the baseline and Top-N configurations, affirming that FLToP CTC surpasses conventional approaches in memory efficiency and aligns with previous WER findings.


\graphicspath{{./images_pdf/}}

\section{Conclusion}
\label{sec:conclusion}

\noindent This study has extensively explored the efficiency and effectiveness of FLToP CTC within ASR systems, specifically analyzing its impact on beam search decoding. Our experiments demonstrate that FLToP CTC significantly reduces the computational load and memory requirements without compromising the transcription accuracy. We found that setting the $N = 4$ effectively balances performance with computational efficiency, evidenced by a significant reduction in beam counts without compromising WER. Our method notably halved the average beam count compared to the baseline and Top-N configurations. Using a dynamic pruning threshold of $R=0.007$ further optimized decoding times while preserving competitive WER results.

\noindent The consistent outperformance of FLToP CTC across various metrics suggests that this approach could serve as a new standard for CTC based ASR decoding, particularly in environments where resource constraints are critical. Future work will focus on refining the adaptive capabilities of the pruning algorithm to enhance its applicability to more diverse and challenging ASR scenarios. Our findings point to a promising direction for future research in ASR technologies, emphasizing the importance of efficiency in decoding strategies to accommodate the growing demand for faster, more accurate ASR systems.

\bibliographystyle{IEEEbib}
\bibliography{strings}

\end{document}